\documentclass[10pt,twocolumn,letterpaper]{article}

\usepackage[pagenumbers]{cvpr}
\usepackage{comment}

\definecolor{cvprblue}{rgb}{0.21,0.49,0.74}

\usepackage[pagebackref,breaklinks,colorlinks,citecolor=cvprblue,linkcolor=red]{hyperref}

\def\paperID{0000}
\def\confName{0000}
\def\confYear{0000}

\usepackage{graphicx}
\usepackage{xspace}
\usepackage{enumitem}
\usepackage{listings}
\usepackage{caption}
\usepackage{amsmath}
\usepackage{amssymb}
\usepackage{multirow}
\usepackage{colortbl}
\usepackage{hhline} 
\usepackage{pifont}
\usepackage{soul}
\usepackage{bbm}
\usepackage{dsfont}
\usepackage{tikz}
\usepackage{pifont}
\usepackage{algorithm} 
\usepackage{algpseudocode} 
\definecolor{darkblue}{rgb}{0, 0.2, 0.6}
\definecolor{darkgray}{rgb}{0.1, 0.1, 0.1}
\definecolor{gray}{rgb}{0.2, 0.2, 0.2}
\definecolor{lightgray}{rgb}{0.3, 0.3, 0.3}
\definecolor{deepdarkblue}{rgb}{0, 0.05, 0.3}
\definecolor{orange}{rgb}{1.0, 0.5, 0.0}
\definecolor{darkorange}{rgb}{0.7, 0.3, 0.0}
\definecolor{realred}{rgb}{1.0, 0.0, 0.0}
\definecolor{pastelred}{rgb}{0.9, 0.2, 0.45}
\definecolor{purple}{rgb}{0.6, 0.0, 0.6}
\definecolor{dogwoodrose}{rgb}{0.84, 0.09, 0.41}
\definecolor{mint}{rgb}{0.01176, 0.5490, 0.5490}
\definecolor{blue}{rgb}{0, 0, 1.0}
\definecolor{azure(colorwheel)}{rgb}{0.0, 0.5, 1.0}
\definecolor{nicegreen}{rgb}{0.0, 0.7, 0.1}
\definecolor{CuGray}{gray}{0.9}
\definecolor{amethyst}{rgb}{0.6, 0.4, 0.8}
\definecolor{black}{rgb}{0.0, 0.0, 0.0}
\definecolor{steelblue}{rgb}{0.27, 0.51, 0.7}
\definecolor{brightcerulean}{rgb}{0.11, 0.67, 0.84}
\definecolor{brown}{rgb}{0.4, 0.3, 0.3}
\definecolor{mainblue}{rgb}{0.12,0.49,0.85}

\newcommand{\commentcolor}[1]{{\color{lightgray}{#1}}}

\makeatletter
\DeclareRobustCommand\onedot{\futurelet\@let@token\@onedot}
\def\@onedot{\ifx\@let@token.\else.\null\fi\xspace}

\def\eg{\emph{e.g}\onedot} 

\def\ie{\emph{i.e}\onedot}

\newcommand*\bigcdot{\mathpalette\bigcdot@{.375}}
\newcommand*\bigcdot@[2]{\mathbin{\vcenter{\hbox{\scalebox{#2}{$\m@th#1\bullet$}}}}}
\usetikzlibrary{tikzmark}
\usetikzlibrary{calc}
\newcommand{\ALGtikzmarkcolor}{black}
\newcommand{\ALGtikzmarkextraindent}{3.5pt}
\newcommand{\ALGtikzmarkverticaloffsetstart}{-1.0ex}
\newcommand{\ALGtikzmarkverticaloffsetend}{-0.5ex}
\newcounter{ALG@tikzmark@tempcnta}
\newcommand\ALG@tikzmark@start{%
    \global\let\ALG@tikzmark@last\ALG@tikzmark@starttext%
    \expandafter\edef\csname ALG@tikzmark@\theALG@nested\endcsname{\theALG@tikzmark@tempcnta}%
    \tikzmark{ALG@tikzmark@start@\csname ALG@tikzmark@\theALG@nested\endcsname}%
    \addtocounter{ALG@tikzmark@tempcnta}{1}%
}
\def\ALG@tikzmark@starttext{start}
\newcommand\ALG@tikzmark@end{%
    \ifx\ALG@tikzmark@last\ALG@tikzmark@starttext
    \else
        \tikzmark{ALG@tikzmark@end@\csname ALG@tikzmark@\theALG@nested\endcsname}%
        \tikz[overlay,remember picture] \draw[\ALGtikzmarkcolor] let \p{S}=($(pic cs:ALG@tikzmark@start@\csname ALG@tikzmark@\theALG@nested\endcsname)+(\ALGtikzmarkextraindent,\ALGtikzmarkverticaloffsetstart)$), \p{E}=($(pic cs:ALG@tikzmark@end@\csname ALG@tikzmark@\theALG@nested\endcsname)+(\ALGtikzmarkextraindent,\ALGtikzmarkverticaloffsetend)$) in (\x{S},\y{S})--(\x{S},\y{E});%
    \fi
    \gdef\ALG@tikzmark@last{end}%
}
\apptocmd{\ALG@beginblock}{\ALG@tikzmark@start}{}{\errmessage{failed to patch}}
\pretocmd{\ALG@endblock}{\ALG@tikzmark@end}{}{\errmessage{failed to patch}}
\makeatother

\renewcommand{\paragraph}[1]{\noindent\textbf{#1.}\,}

\newcommand{\institutiontext}[1]{{\fontsize{10}{12}\selectfont #1}}

\newcommand{\LineNumberAlg}[1]{\makebox[1.36em][r]{#1\;\!:}}
\newcommand{\refclr}[1]{{\color{red}{#1}}}

\title{Robust 3D Shape Reconstruction in Zero-Shot from a Single Image in the Wild}
\author{%
\vspace{1mm}
{Junhyeong Cho${}^1$ \quad
Kim Youwang${}^2$ \quad
Hunmin Yang${}^{1,3}$ \quad
Tae-Hyun Oh${}^{2,4}$}\\
\vspace{-0.5mm}
\institutiontext{%
${}^1$ADD \quad
${}^2$Department of Electrical Engineering, POSTECH}\\ 
\institutiontext{%
${}^3$Department of Mechanical Engineering, KAIST \quad 
${}^4$School of Computing, KAIST}\\
\institutiontext{%
\url{https://ZeroShape-W.github.io}}%
}

\begin{document}

\maketitle

\begin{abstract}
Recent monocular 3D shape reconstruction methods have shown promising zero-shot results on object-segmented images without any occlusions. However, their effectiveness is significantly compromised in real-world conditions, due to imperfect object segmentation by off-the-shelf models and the prevalence of occlusions. 
To effectively address these issues, we propose \textbf{a unified regression model} that integrates segmentation and reconstruction, specifically designed for occlusion-aware 3D shape reconstruction. 
To facilitate its reconstruction in the wild, we also introduce \textbf{a scalable data synthesis pipeline} that simulates a wide range of variations in objects, occluders, and backgrounds. Training on our synthetic data enables the proposed model to achieve state-of-the-art zero-shot results on real-world images, using significantly fewer parameters than competing approaches.
\end{abstract}
\vspace{-1mm}
\section{Introduction}
\begin{figure}[t!]
    \centering
    \includegraphics[width=0.999\columnwidth]{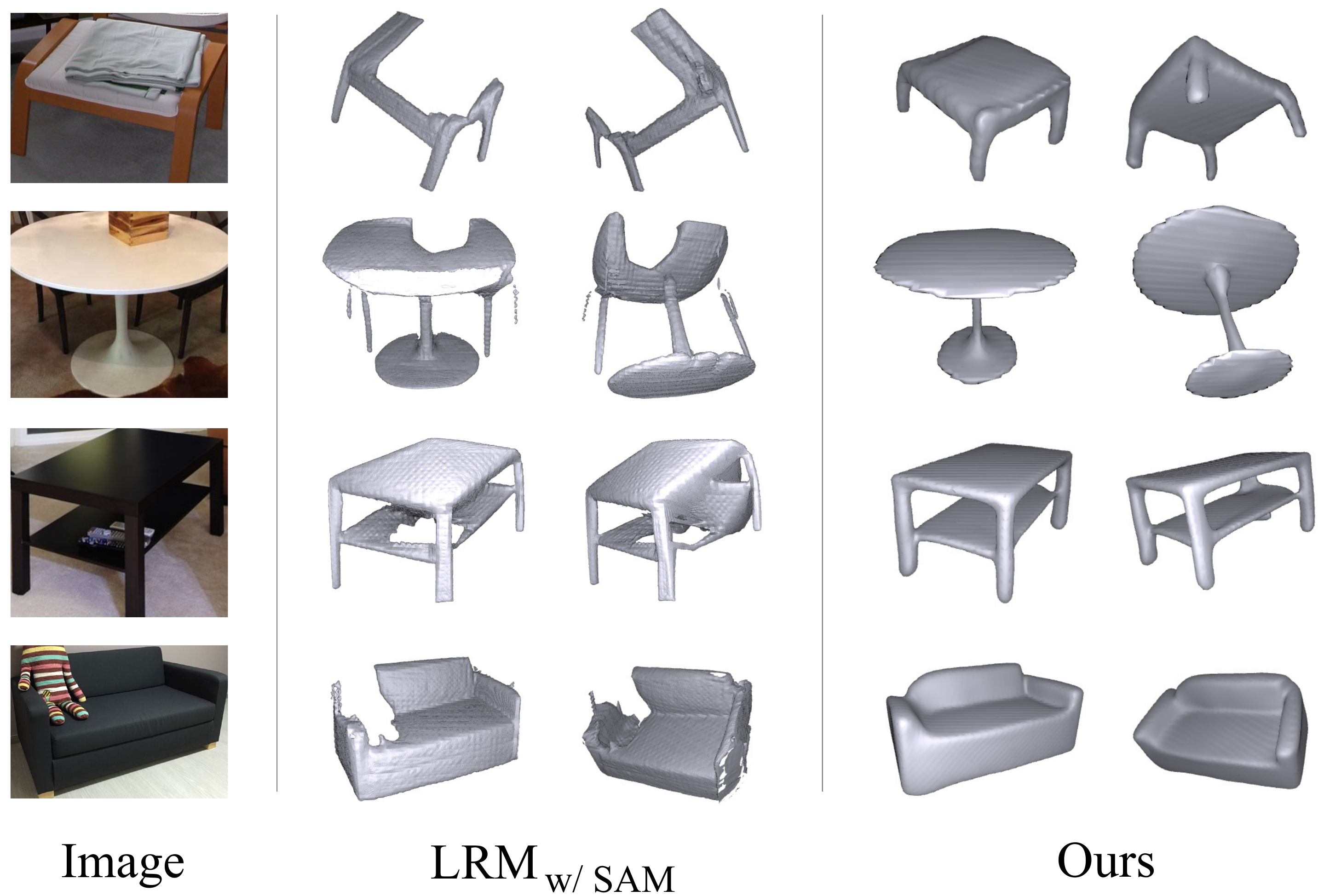}
    \vspace{-6.6mm}
    \caption{Single-view 3D shape reconstruction on real-world images~\cite{Pix3D}.
    Compared with LRM~\cite{hong2024lrm} that requires an off-the-shelf segmentation model (\eg, SAM~\cite{SAM}),
    our approach not only eliminates the need for such an additional model but also achieves greater robustness to occlusions.}
    \label{fig:fig_intro_teaser}
    \vspace{-1.75mm}
\end{figure}

The abundance of large-scale 3D shapes~\cite{ShapeNet,ABO,ABC,Objaverse-XL,3D-FUTURE} has enabled the development of reconstruction models with strong generalization capabilities, even allowing for zero-shot monocular 3D shape reconstruction~\cite{hong2024lrm,huang2024zeroshape,Wonder3D,LGM,TripoSR,CRM}. However, these models are not deployable in real-world settings, as they assume object images that have no backgrounds and occlusions. 
Thus, they utilize \textit{off-the-shelf} segmentation models~\cite{SAM,SAM2} to handle in-the-wild images.
This dependency introduces two main challenges: (i) error accumulation by off-the-shelf models, and (ii) removal of occluded object parts by  segmentation.
Both issues hinder precise shape reconstruction, as shown in Figure~\ref{fig:fig_intro_teaser}.

\begin{figure*}[t!]
    \centering
    \includegraphics[width=0.995\textwidth]{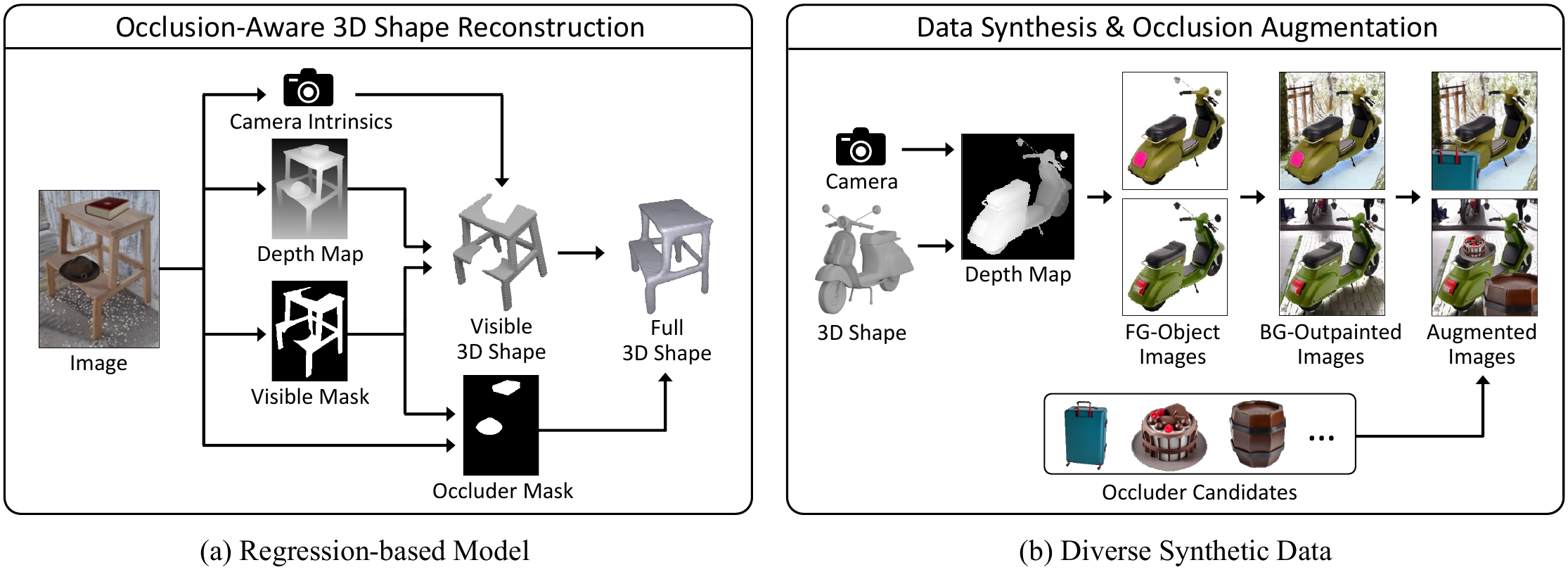}
    \vspace{-2.75mm}
    \caption{Proposed approach for zero-shot monocular 3D shape reconstruction in the wild.
    (a) Our model reconstructs the full 3D shape of a salient object using its visible 3D shape and the silhouette of its occluders. 
    The visible 3D shape is estimated from camera intrinsics, depth map, and visible region of the object.
    (b) We create our training dataset by synthesizing diverse data.
    We render 3D shapes and then simulate their appearances and backgrounds using generative models. Occluders are inserted on-the-fly during model training.}
    \label{fig:fig_intro_overview}
    \vspace{-1.5mm}
\end{figure*}

To tackle the challenges, we propose \textbf{ZeroShape-W}, an occlusion-aware 3D shape reconstruction model (Fig.~\ref{fig:fig_intro_overview}\refclr{a}).
Our model jointly regresses the silhouettes of a salient object and its occluders, along with a depth map and camera intrinsics to estimate the object's visible 3D shape.
Then, its full 3D shape is reconstructed by using the visible 3D shape and the silhouette of occluders.
This approach enables the reconstruction of occluded object parts, making it effective in real-world conditions. 
However, a significant challenge remains: \textit{how can we train our model for robust zero-shot reconstruction in the wild?}
To achieve this goal, we need large-scale diverse data that can cover various real environments.
Unfortunately, the availability of real-world images annotated with precise 3D shapes is limited. 
An alternative approach is to generate realistic renderings from 3D object collections~\cite{Objaverse,Objaverse-XL,3D-FUTURE,ABO,ShapeNet}, but this is also limited by the availability of high-quality synthetic assets.

In this paper, we introduce a scalable data synthesis pipeline (Fig.~\ref{fig:fig_intro_overview}\refclr{b}) that simulates a wide range of variations using generative models, thereby removing the reliance on limited real-world data or high-quality synthetic assets. Our pipeline first synthesizes foreground object images using a conditional generative model~\cite{ControlNet}, with spatial conditions obtained by rendering 3D shapes available from extensive object collections~\cite{ShapeNet,Objaverse-XL,Objaverse}. Then, their backgrounds are generated by an object-aware background outpainting model~\cite{salientobjbg}. For these synthesized images, we apply Copy-Paste augmentation~\cite{CopyPaste} to simulate occluders on-the-fly during model training.

By training the proposed model on our diverse synthetic data,
we expose it to a wide variety of foreground objects, their occluders, and backgrounds. This process enables our model to focus on learning to capture domain-invariant and object-centric geometric priors. In this way, we improve its generalization across various environments.

To validate the effectiveness of our approach, 
we first create our training dataset by synthesizing diverse images using 3D shapes from Objaverse~\cite{Objaverse} and ShapeNet~\cite{ShapeNet}. Then, we train our model on the data augmented with occlusions.
We evaluate the model on the real-world benchmark, Pix3D~\cite{Pix3D},
where it outperforms state-of-the-art models while using significantly fewer model parameters.
In addition to the benchmark, we also qualitatively evaluate the proposed model on in-the-wild images from ObjectNet~\cite{Objectnet}, OfficeHome~\cite{OfficeHomedataset}, and PACS~\cite{PACSdataset}. 

\section{Related Work}

\paragraph{Single-view 3D shape reconstruction in the wild}
This task aims to estimate the 3D shape of an object from a single image captured in a real-world environment. It is challenging because objects exhibit a variety of 3D shapes while their appearances are greatly altered by environmental factors. To handle such variations, 3D shape reconstruction models~\cite{GenRe,TMN,mescheder2019occupancy,LDIF,pixel2mesh,3DGAN,3D-R2N2,MarrNet,ShapeHD,meshrcnn,Total3D,im3D,instPIFu,girdhar2016learning,tulsiani2017multi,AtlasNet,alwala2022pretrain,kanazawa2018learning,goel2020shape,huang2023shapeclipper,hong2024lrm,huang2024zeroshape,Wonder3D,LGM,TripoSR,CRM,InstantMesh,metta,PT43D} should generalize well across various environments. However, the scarcity of real-world 3D data poses a significant obstacle in pursuing this goal. In this paper, we address this issue by synthesizing large-scale data using generative models.

Single-view reconstruction methods fall into two categories: generative model-based approaches~\cite{Wonder3D,LGM,CRM,InstantMesh} and regression-based approaches~\cite{huang2024zeroshape,TripoSR,hong2024lrm}.
The generative model-based approaches synthesize multi-view images using diffusion models such as Zero-1-to-3~\cite{liu2023zero1to3},
which are effective to estimate 3D shapes with textures but incur significant computational costs.
In contrast, regression-based approaches such as ZeroShape~\cite{huang2024zeroshape} estimate 3D shapes in a single forward pass, offering much greater efficiency.

Recent state-of-the-art methods in both approaches assume clean object-segmented images without occlusions, 
making them difficult to operate in real-world settings.
In this work, we tackle this issue by introducing a unified regression model that both segments and reconstructs while accounting for occlusions.
To the best of our knowledge, our work is the first regression-based approach for zero-shot single-view 3D shape reconstruction in the wild.

\begin{figure*}[t!]
    \centering
    \includegraphics[width=0.999\textwidth]{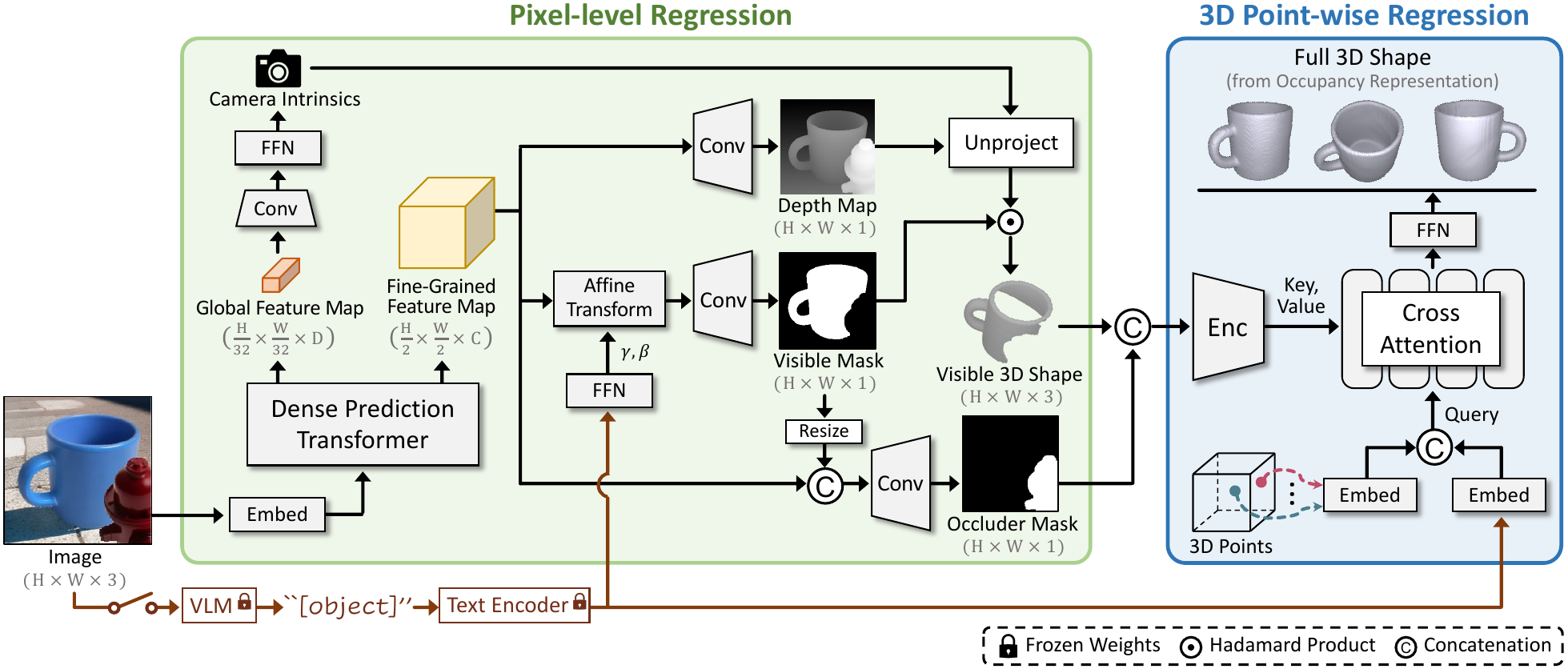}
    \vspace{-5.75mm}
    \caption{Overall architecture of ZeroShape-W. Given an object-centric RGB image, our model leverages the backbone of Dense Prediction Transformer (DPT)~\cite{DPT} to regress camera intrinsics, 
    a depth map, a visible mask of a salient object, and an occluder mask that represents the object's occluders.
    These components are used to derive the object's visible 3D shape, which is then combined with the occluder mask to regress occupancy values of 3D point queries through cross-attention layers~\cite{vaswani2017attention}. 
    This process recovers the object's full 3D shape, including occluded object parts.
    To alleviate difficulties of in-the-wild reconstruction, 
    we optionally incorporate open-set category priors by estimating the object's category, ``\texttt{[object]}'' (\eg, ``cup''), using an additional vision-language model (VLM)~\cite{LLaMA-VID}.}
    \label{fig:fig_method}
    \vspace{-0.5mm}
\end{figure*}

\paragraph{Improving model robustness to distribution shifts} 
It is common to exploit multiple domains for learning domain-invariant features~\cite{Zhou2020Learning,Li2019Episodic,Carlucci2019Domain,Dou2019Domain,Matsuura2020Domain,Seo2020Learning,Mahajan2021Domain,CDS,zhou2021mixstyle,koh2021wilds} to enhance the model robustness.
However, multiple domains from real-world environments 
might not be accessible. To effectively tackle this issue, domain randomization methods~\cite{DR,DetectionDR,PoseDR,FSDR,DR-PC,DR-workshop,Structured-DR,Active-DR,exarchos2021policy} synthesize multiple domains via a random simulation of diverse scenarios.
By exposing a model to randomized variations, it learns to capture generalizable representations capable of handling a wide range of variations encountered in real-world settings.

Following the paradigm of domain randomization,
we simulate diverse variations in foreground objects, occluders, and backgrounds. However, this is not done completely randomly,
as the background outpainting model~\cite{salientobjbg} generates backgrounds while considering the foreground objects.
By training the proposed model on our data, it could learn to capture domain-invariant and object-centric geometric priors, which are useful for handling diverse objects.

\paragraph{Synthetic data generation}
To train 3D shape reconstruction models, we can synthesize data by rendering 3D objects from extensive object collections~\cite{ShapeNet,ABO,ABC,Objaverse-XL,Objaverse,3D-FUTURE} via computer graphics tools.
As long as sufficient 3D shape and texture assets are available for objects and environments, this approach could also facilitate 3D shape reconstruction in the wild.
However, it is expensive to collect a great variety of high-quality assets (\eg, 4K-resolution texture maps, HDR environment maps) needed to bridge the gap between synthetic and real environments. 
Moreover, the diversity of object appearances and backgrounds depicted by rendered images is determined by the combination of available assets.
Compared with these computer graphics tools, our data synthesis pipeline could simulate more diverse object appearances and backgrounds without relying on such additional high-quality assets, by leveraging generative models pre-trained on large-scale datasets.

\paragraph{Generative models with 3D assets}
Recent works~\cite{ma2024generating,ge20243d,zhang2024hoidiffusion} use conditional generative models~\cite{ControlNet,T2I-Adapter,UniControl} to synthesize images with 3D assets for various tasks, \eg, object pose estimation. Compared with them, accurate preservation of silhouettes within spatial conditions is more crucial for 3D shape reconstruction. To better preserve silhouettes, our data synthesis pipeline employs a two-step generation process. We utilize an initial guidance for foreground object synthesis and exploit a background outpainting model~\cite{salientobjbg} specifically designed for preserving silhouettes. In this way, each synthesized image is paired with its precise 3D shape.

\begin{figure}[t!]
    \centering
    \includegraphics[width=0.99\columnwidth]{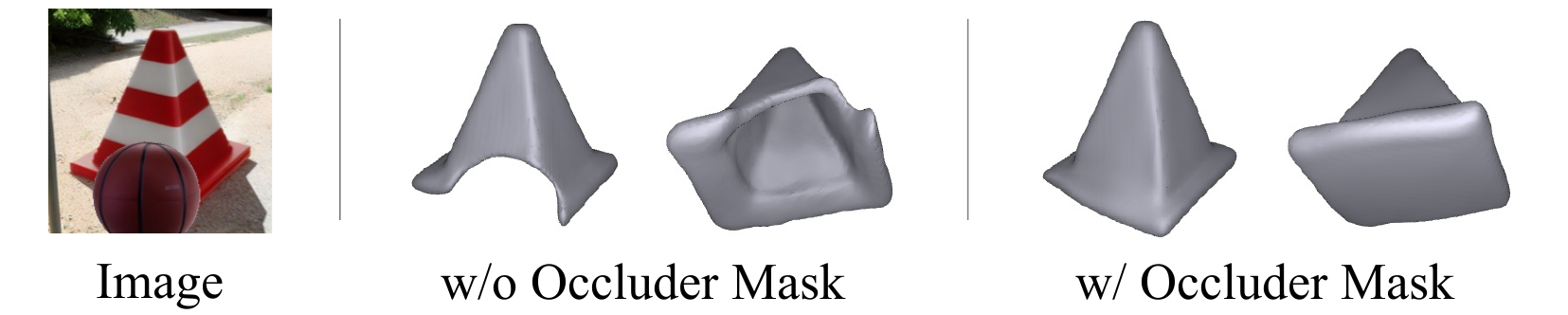}
    \vspace{-1.5mm}
    \caption{Effect of using an occluder mask. It enables full 3D shape reconstruction, even for occluded object parts.
    Note that existing state-of-the-art methods~\cite{hong2024lrm,huang2024zeroshape,Wonder3D,LGM,TripoSR,CRM} struggle with occlusions, as they assume the salient visible object in the current view is unobstructed.}
    \label{fig:fig_occ}
    \vspace{-3.5mm}
\end{figure} 

\section{Method}
\label{sec:method}

The proposed ZeroShape-W
(Fig.~\ref{fig:fig_method}) aims to regress the full 3D shape of a salient object from a single image, including its visible and occluded parts (Fig.~\ref{fig:fig_occ}). 
Inspired by~\cite{huang2024zeroshape,MCC}, our model first regresses pixel-level information, which is used as condition to regress occupancy values of 3D point queries. These occupancy values represent the visible and hidden geometry of the salient object, while also accounting for occluded regions that were not addressed in~\cite{huang2024zeroshape,MCC}.

During pixel-level regression, our model regresses a depth map, a visible mask of the object, and an occluder mask that depicts the object's occluders. It also estimates camera intrinsics to convert depth values into the object's visible 3D shape in a viewer-centric coordinate system.

During 3D point-wise regression, our model regresses occupancy values of 3D point queries through cross-attention layers~\cite{vaswani2017attention}, utilizing features from the visible 3D shape and occluder mask as keys and values.
This approach can effectively reconstruct the object's hidden geometry, with the help of learned 3D shape priors.

For in-the-wild images, identifying arbitrary salient objects is often challenging due to complex backgrounds or occlusions, which can result in noisy mask estimations.
In such cases, visible 3D shapes may mistakenly include background geometries.
To address these issues, we optionally incorporate category-specific priors by leveraging a vision-language model~\cite{LLaMA-VID} and CLIP text encoder~\cite{radford2021clip}.
Specifically, the vision-language model generates a \textbf{category-specific prompt} ``\texttt{[object]}'' (\eg, ``cup'' in Fig.~\ref{fig:fig_method}) that describes the salient object in the input image.
For stable inference with unseen object categories, we use an averaged CLIP text embedding obtained from the category-specific prompt and a \textbf{category-agnostic prompt} ``{object}''.
In cases where the category-specific priors are not used, our model relies solely on the category-agnostic prompt without estimating the object's category.

\subsection{Pixel-level Regression}
\label{sec:subsec_pixel_level_regression}

For an object-centric RGB image $I \in \mathbb{R}^{H \times W \times 3}$, we first obtain image embeddings $\mathbf{E}_I \in \mathbb{R}^{(H/16) \times (W/16) \times D}$ using ResNet50~\cite{resnet}, as done in ViT-Hybrid~\cite{dosovitskiy2021an}. Then, we feed the embeddings $\mathbf{E}_I$ into the backbone of Dense Prediction Transformer (DPT)~\cite{DPT}.
From the backbone, we extract a global feature map $\mathbf{X}_G \in \mathbb{R}^{(H/32) \times (W/32) \times D}$ and a fine-grained feature map $\mathbf{X}_F \in \mathbb{R}^{(H/2) \times (W/2) \times C}$. The global feature map $\mathbf{X}_G$ is the reassembled output from the last transformer layer in DPT, while the fine-grained feature map $\mathbf{X}_F$ 
is the output from the last fusion layer in DPT.

\paragraph{Camera intrinsics}
To regress the visible 3D shape of the salient object, we estimate camera intrinsics $K \in \mathbb{R}^{3 \times 3}$ from the global feature map $\mathbf{X}_G$. Specifically, we obtain the camera intrinsics $K$ by predicting a focal length scaling factor and principal point shifts, as done in~\cite{huang2024zeroshape}. The scaling factor adjusts a base focal length according to the image resolution $H \times W$, and the principal point shifts are also specified relative to the image resolution. These parameters are estimated from the global feature map $\mathbf{X}_G$ using shallow convolutional layers, a pooling layer, and a feed forward network (FFN).

\paragraph{Depth map}
To derive the visible 3D shape by unprojecting depth values using the estimated camera intrinsics $K$,
we regress a depth map $\mathbf{M}_D \in \mathbb{R}^{H \times W \times 1}$ from the fine-grained feature map $\mathbf{X}_F$.
For this regression, we use shallow convolutional layers with interpolation.

\paragraph{Visible mask}
To identify the visible region of the salient object, we regress a visible mask $\mathbf{M}_V \in \mathbb{R}^{H \times W \times 1}$ from the fine-grained feature map $\mathbf{X}_F$.
For precise regression, we adaptively adjust the fine-grained feature map $\mathbf{X}_F$ based on the object's category. Inspired by~\cite{hong2024lrm,DiT,FiLM}, we modulate the feature map using an affine transformation with a scaling factor $\gamma \in \mathbb{R}^{C}$ and a shift factor $\beta \in \mathbb{R}^{C}$, estimated by an FFN with CLIP text encoder~\cite{radford2021clip}. We modulate the feature map $\mathbf{X}_F$ at each spatial location $(i,j)$ as follows:
\begingroup
\setlength{\abovedisplayskip}{5pt}
\setlength{\belowdisplayskip}{5pt}
    \begin{align}
        \label{eq:feat_mod}
        \mathbf{\bar{X}}_{F_{ij}} = (1+\gamma)\mathbf{X}_{F_{ij}} + \beta,
    \end{align}
\endgroup
\noindent where $\mathbf{\bar{X}}_F \in \mathbb{R}^{(H/2) \times (W/2) \times C}$. From this modulated feature map,
we regress the visible mask $\mathbf{M}_V$ using shallow convolutional layers with interpolation.

\paragraph{Visible 3D shape}
Using the estimated camera intrinsics $K$, depth map $\mathbf{M}_D$, and visible mask $\mathbf{M}_V$, we derive the visible 3D shape $\mathbf{S}_V \in \mathbb{R}^{H \times W \times 3}$ that contains $(x,y,z)$-coordinates of each pixel location $(i,j)$ as follows:
\begingroup
\setlength{\abovedisplayskip}{5pt}
\setlength{\belowdisplayskip}{5pt}
    \begin{align}
        \label{eq:unproject}
        \mathbf{S}_{V_{ij}} = 
        \mathds{1}_{\{ \mathbf{M}_{V_{ij}} \geq \eta \}}
        \bigcdot
        (\mathbf{M}_{D_{ij}} K^{-1}[i,j,1]^{\top}),
    \end{align}
\endgroup
where $\mathds{1}_{\{\cdot\}}$ is an indicator function, and $\eta$ denotes a threshold.
We normalize $\mathbf{S}_V$ to have a zero mean and a unit scale.

\paragraph{Occluder mask}
To determine the object's occluded region, we regress an occluder mask $\mathbf{M}_O \in \mathbb{R}^{H \times W \times 1}$ using the fine-grained feature map $\mathbf{X}_F$ and visible mask $\mathbf{M}_V$. Specifically, we resize the visible mask to the resolution of $H/2 \times W/2$, and concatenate it with the feature map along the channel dimension. Then, we regress the occluder mask 
$\mathbf{M}_O$ using shallow convolutional layers with interpolation.

\subsection{3D Point-wise Regression}
\label{sec:subsec_3D_point_wise_regression}

We regress occupancy values of 3D point queries using cross-attention layers to effectively consider local features from pixel-level estimations.\footnote{We extend prior works~\cite{MCC, huang2024zeroshape} to reconstruct occluded object parts.}
We begin by concatenating the visible 3D shape $\mathbf{S}_V$ and the occluder mask $\mathbf{M}_O$ along the channel dimension, which is then fed into an encoder.
It produces $Z$-dimensional vectors that serve as keys and values in cross-attention layers.

For each 3D point query, we construct a $Z$-dimensional query vector by concatenating two $Z/2$-dimensional embeddings: one from the point's $(x,y,z)$-coordinates and the other from CLIP text encoder~\cite{radford2021clip}. The query vectors are processed through $L$ cross-attention layers, where each query independently attends to its relevant spatial features. Then, to reconstruct the object's full 3D shape,
we estimate the occupancy value of each query using an FFN.

\subsection{Loss}

We train our model using binary cross-entropy losses with the estimated visible mask $\mathbf{M}_V$, occluder mask $\mathbf{M}_O$, and occupancy values.
For the estimated depth map $\mathbf{M}_D$, we apply a scale- and shift-invariant MAE loss~\cite{ranftl2020robustmonoculardepthestimation} over two regions:
(i) the visible mask region with ground-truth depth
and (ii) the entire region with pseudo depth estimated from a pre-trained model, referred to as the auxiliary depth loss.
This auxiliary loss addresses the limitation of our data synthesis pipeline, which only provides depth for object areas.
For the estimated camera intrinsics $K$, we use an MSE loss by comparing the visible 3D shape $\mathbf{S}_V$ to the ground truth.

\begin{figure*}[t!]
    \centering
    \includegraphics[width=0.99\textwidth]{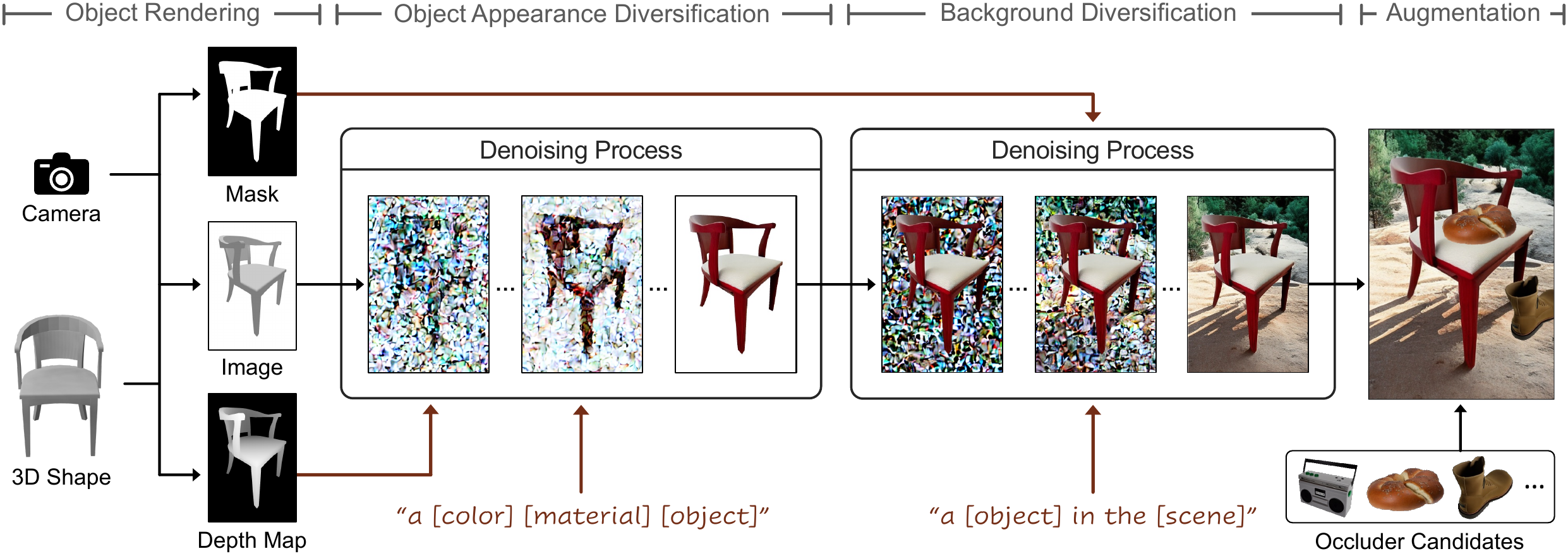}
    \vspace{-1.75mm}
    \caption{Overview of our data synthesis. Given a camera and 3D object, we render it to obtain an object mask, image, and depth map. We simulate object appearances via a conditional diffusion model~\cite{ControlNet} using the depth as its spatial condition and ``a \texttt{[color]} \texttt{[material]} \texttt{[object]}'' (\eg, ``a red wood chair'') as its textual condition.
    To alleviate shape distortion by the generative model, we use the rendered image as initial guidance.
    We simulate its background via an object-aware background outpainting model~\cite{salientobjbg} using the mask and ``a \texttt{[object]} in the \texttt{[scene]}'' (\eg, ``a chair in the canyon'') as its textual condition. Then, we put occluders during data augmentation.}
    \label{fig:fig_data_syn}
    \vspace{-3mm}
\end{figure*} 

\subsection{Implementation Details}
\label{subsec:impl_details}

We input an image of resolution $H$$\,\times\,$$W$, where $H=W=224$. Using DPT-Hybrid~\cite{DPT}, we extract the $D$-dimensional feature $\mathbf{X}_G$ and $C$-dimensional feature $\mathbf{X}_F$, where $D$$\,=\,$$768$ and $C$$\,=\,$$256$. In eq.~(\ref{eq:unproject}), we set $\eta$$\,=\,$$0.5$. In $L$ cross-attention layers, we employ $Z$-dimensional vectors, where $L$$\,=\,$$2$ and $Z$$\,=\,$$256$. During training, we use Depth\,Anything\,V2~\cite{DAV2} for the auxiliary depth loss. At inference time, we optionally use LLaMA-VID~\cite{LLaMA-VID} to identify the object's category. Further details can be found in the supplementary material.
\section{Data Synthesis}

Our scalable data synthesis pipeline (Fig.~\ref{fig:fig_data_syn}) produces diverse images, along with their corresponding 3D shapes, camera parameters, depth maps, and masks.
We use Copy-Paste~\cite{CopyPaste} augmentation to insert occluders at each training iteration. Please see the supplementary material for details.

\subsection{Object Rendering}
\label{sec:subsec_rendering}

A great variety of 3D shapes can be readily obtained from 3D object collections~\cite{ShapeNet,ABO,ABC,Objaverse-XL,Objaverse,3D-FUTURE} or generative models~\cite{MeshDiffusion,3DGAN,Diffusion-SDF,SDF-Diffusion}.
In line with ZeroShape~\cite{huang2024zeroshape}, we utilize the same set of 3D objects spanning over 1,000 categories, including 52K objects from ShapeNetCore.v2~\cite{ShapeNet} and 42K objects from Objaverse-LVIS~\cite{Objaverse}.

Each 3D object is rendered by multiple cameras constructed with varying camera distances, elevation angles, focal lengths, and LookAt points.
This rendering process produces over 1 million object images,
and each output is paired with precise 3D shapes, camera parameters, depth maps, and masks.

\subsection{Object Appearance Diversification}
\label{sec:subsec_appearance}

To simulate diverse visual variations in object appearances, 
we employ ControlNet~\cite{ControlNet} which is a conditional diffusion model capable of realistically synthesizing diverse images.
In our data synthesis, the generative model simulates a wide range of variations by using a depth map as its spatial condition and ``a \texttt{[color][material][object]}'' as its textual condition.
Here, the placeholders \texttt{[color]} and \texttt{[material]} are replaced by words selected from pre-defined color and material lists, while \texttt{[object]} denotes the category of the used 3D object.

\begin{figure}[t!]
    \centering
    \includegraphics[width=0.95\columnwidth]{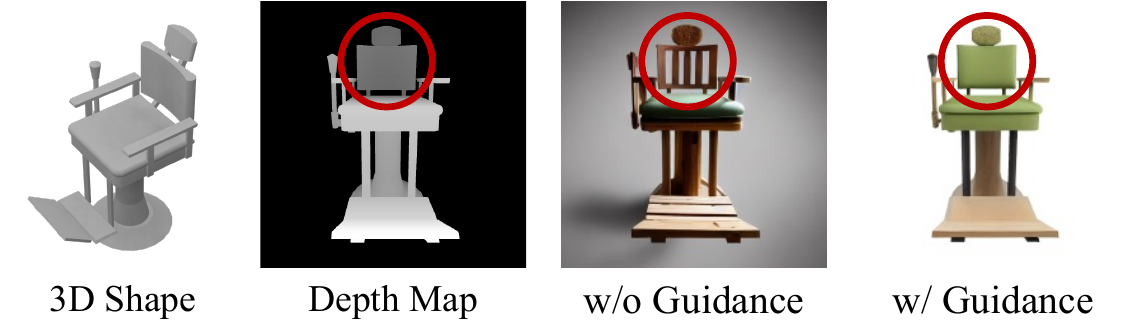}
    \vspace{-2.25mm}
    \caption{
    Effect of using an initial guidance. 
    The guidance assists in preserving the object silhouette depicted in the spatial condition.}
    \label{fig:fig_guidance}
    \vspace{-3.75mm}
\end{figure}

\paragraph{Initial guidance}
The diffusion model~\cite{ControlNet} occasionally distorts the object silhouette depicted in the spatial condition.
To alleviate this distortion, we utilize the rendered image as initial guidance. Motivated by SDEdit~\cite{SDEdit},
we perturb the guidance with Gaussian noise and feed it to the model.
Since the noisy guidance retains the original spatial structure, it effectively reduces silhouette distortion (Fig.~\ref{fig:fig_guidance}).

\begin{figure*}[t!]
    \centering
    \includegraphics[width=0.999\textwidth]{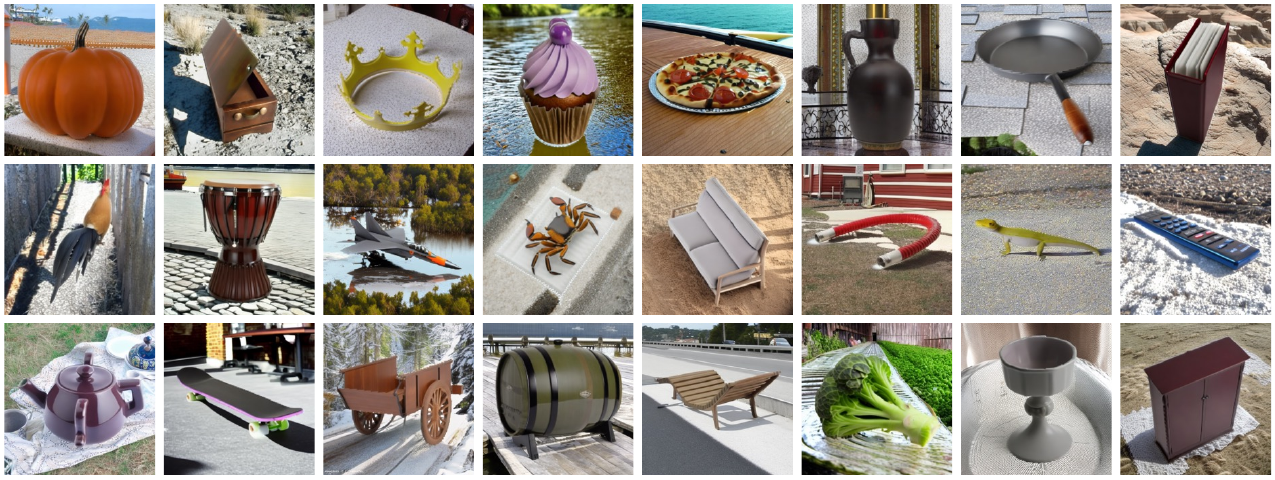}
    \vspace{-6mm}
    \caption{Diverse synthetic data produced by our scalable data synthesis pipeline. 
    Based on 3D shape renderings from ShapeNetCore.v2~\cite{ShapeNet} and Objaverse-LVIS~\cite{Objaverse}, we synthesize diverse images using ControlNet~\cite{ControlNet} and object-aware background outpainting model~\cite{salientobjbg}.}
    \label{fig:fig_data_examples}
\end{figure*}

\subsection{Background Diversification}
\label{sec:subsec_background}

To simulate diverse backgrounds, we use an object-aware background outpainting model~\cite{salientobjbg}. It adapts Stable Diffusion Inpainting 2.0~\cite{StableDiffusion} by using modified ControlNet~\cite{ControlNet},
allowing for background generation without extending the object's silhouette.
We generate various backgrounds using the object image, its mask, and the prompt ``a \texttt{[object]} in the \texttt{[scene]}'', where \texttt{[scene]} is selected from scene categories~\cite{SUN,Places365}. Ideally, it synthesizes backgrounds that appear to be captured by cameras with focal lengths similar to those used for the object’s rendering.

\subsection{Augmentation}

For images generated by our data synthesis (Fig.~\ref{fig:fig_data_examples}), we simulate occluders on-the-fly during model training via Copy-Paste augmentation~\cite{CopyPaste} using our synthesized foreground objects. To enable stable estimation of camera intrinsics, we put objects rendered by similar focal lengths.

\section{Experiments}
\label{sec:experiments}

\subsection{Evaluation Settings}

\paragraph{Protocol} 
To convert implicit 3D shape representation into explicit 3D meshes, we extract the isosurface from the implicit representation via the Marching Cubes~\cite{MarchingCubes} algorithm. Following~\cite{AtlasNet,Total3D,im3D}, we align predicted and ground-truth meshes using the Iterative Closest Point algorithm. Then, we uniformly sample 10K points from each mesh.

\paragraph{Metrics} Using 10K sampled points, we compute Chamfer Distance (CD) and F-Score (FS) for quantitative evaluation, as in~\cite{huang2024zeroshape,huang2022planesvschairscategoryguided,huang2023shapeclipper}.
Specifically, FS@$\tau$ denotes the harmonic mean of precision@$\tau$ and recall@$\tau$,
which represents the concordance between predicted and ground-truth points within a distance threshold $\tau$.

\begin{table*}[!t]
    \centering
    \resizebox{\textwidth}{!}{
        \begin{tabular}{l c c c c c c c c c | c}
        \toprule
        & \multicolumn{2}{c}{Off-the-shelf\;\,Model} & & Overhead & & \multicolumn{5}{c}{Pix3D\;\,Evaluation}
        \\
        \cline{2-3}
        \cline{5-5}
        \cline{7-11}
        \vspace{-0.8mm}
        & \raisebox{-0.15ex}{Modal} & \raisebox{-0.15ex}{Amodal} & & & & & & & &
        \\
        \!Model & Segmentation & Completion  & & \#Params & & FS@$\tau$\;\!$\uparrow$ & FS@2$\tau$\;\!$\uparrow$ & FS@3$\tau$\;\!$\uparrow$ & FS@5$\tau$\;\!$\uparrow$ & CD\,$\downarrow$
        \\
        \hline
        \multirow{2}{*}{\!LRM} & SAM & --- & & $>\,$1100M & & 31.0 & 54.5 & 69.9 & 87.1 & 0.121
        \\
        & SAM & pix2gestalt & & $>\,$2400M & & 31.1 & 54.9 & 70.6 & 87.7 & 0.119
        \\
        \hline
        \multirow{2}{*}{\!ZeroShape} & SAM & --- & & $>\,$800M & & 32.1 & 56.8 & 72.1 & 88.0 & 0.116
        \\
        & SAM & pix2gestalt & & $>\,$2100M & & 33.6 & 59.0 & 74.2 & 89.2 & 0.110
        \\
        \hline
       \cellcolor{gray!7.5}\textbf{\!Ours {\small{(category-agnostic)}}}\!\! & \cellcolor{gray!7.5}--- & \cellcolor{gray!7.5}--- & \cellcolor{gray!7.5} & \cellcolor{gray!7.5}\textbf{193.7M} & \cellcolor{gray!7.5} & \cellcolor{gray!7.5}\textbf{38.2} & \cellcolor{gray!7.5}\textbf{65.3} & \cellcolor{gray!7.5}\textbf{79.9} & \cellcolor{gray!7.5}\textbf{92.5} & \cellcolor{gray!7.5}\textbf{0.097}
       \\
       \bottomrule
    \end{tabular}}
    \vspace{-1mm}
    \caption{Comparison of single-view 3D shape reconstruction on Pix3D~\cite{Pix3D}.
    Existing state-of-the-art reconstruction models, LRM~\cite{hong2024lrm} and ZeroShape~\cite{huang2024zeroshape}, require an off-the-shelf modal segmentation model (\eg, SAM~\cite{SAM}), because they assume object-segmented images without any occlusions.
    For occlusion-aware reconstruction, they need an additional amodal completion model (\eg, pix2gestalt~\cite{pix2gestalt}). 
    In this comparison, our model is evaluated using the category-agnostic prompt ``{object}''.}
    \vspace{1.5mm}
    \label{table:table_main_result}
\end{table*}

\begin{figure*}[t!]
    \centering
    \includegraphics[width=0.999\textwidth]{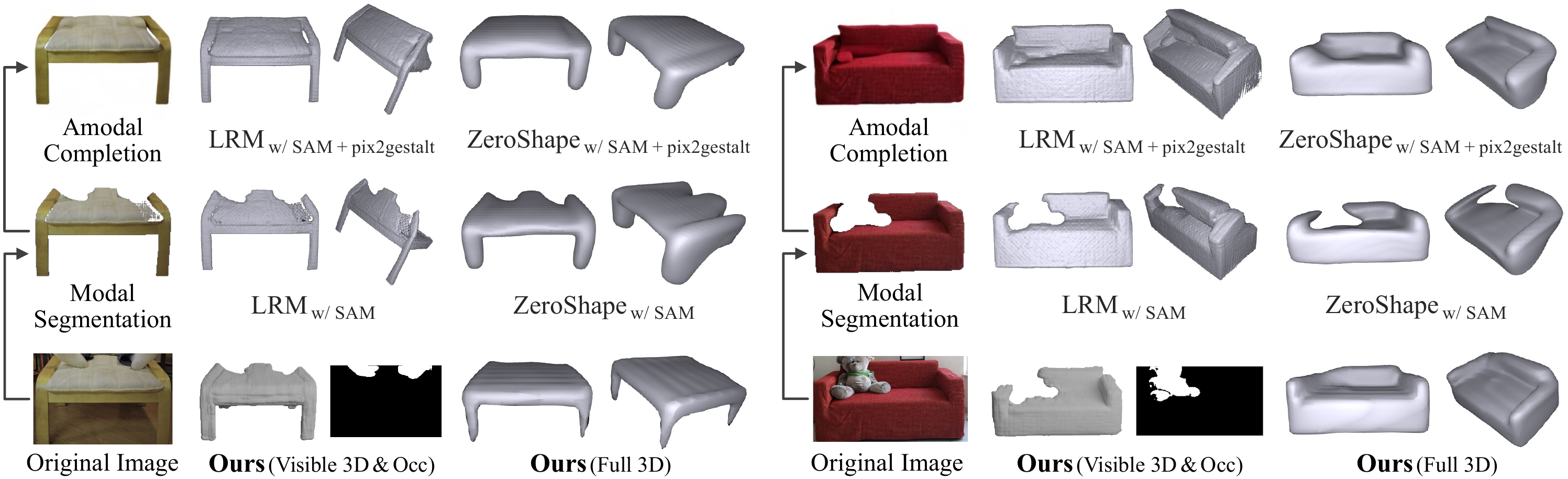}
    \vspace{-5.5mm}
    \caption{Comparison of single-view 3D shape reconstruction on Pix3D~\cite{Pix3D}. 
    In this comparison, LRM~\cite{hong2024lrm} and ZeroShape~\cite{huang2024zeroshape} take modal segmentation results (from SAM~\cite{SAM}) or amodal completion results (from pix2gestalt~\cite{pix2gestalt}) as inputs.
    In contrast, our model directly takes original images as inputs, and performs occlusion-aware reconstruction by regressing visible 3D shapes and occluder silhouettes.%
    }
    \vspace{2mm}
    \label{fig:qualitative_comparison}
\end{figure*}

\paragraph{Datasets} We evaluate our model on real-world object images from Pix3D~\cite{Pix3D}, ObjectNet~\cite{Objectnet}, OfficeHome~\cite{OfficeHomedataset}, and PACS~\cite{PACSdataset}. As Pix3D is the widely recognized benchmark for 3D shape reconstruction in the wild, we use it for comparison with other state-of-the-art models~\cite{hong2024lrm,huang2024zeroshape}.

\subsection{Experimental Results}

\paragraph{Quantitative comparison} 
Currently, there is no zero-shot monocular 3D shape reconstruction model capable of directly processing in-the-wild images. Therefore, we compare our model with state-of-the-art reconstruction models, LRM~\cite{hong2024lrm}\footnote{In this paper, we use the publicly available OpenLRM model weights from \url{https://github.com/3DTopia/OpenLRM}} and ZeroShape~\cite{huang2024zeroshape}, which operate on object-segmented images without occlusions.
For comprehensive comparisons with these baselines, we equip them with the off-the-shelf modal segmentation model, SAM~\cite{SAM}, and amodal completion model, pix2gestalt~\cite{pix2gestalt}.

In Table~\ref{table:table_main_result}, we evaluate the baselines and our model on the Pix3D benchmark~\cite{Pix3D}. Although the proposed model does not utilize such off-the-shelf models, it substantially outperforms them.
In terms of FS@$\tau$, our model outperforms the strongest baseline ZeroShape$_{\mathrm{w/ SAM \,+\, pix2gestalt}}$ by a large margin of 13.7\% (38.2 vs. 33.6).
With regard to CD, the proposed model also surpasses the strongest baseline by a large margin of 13.4\% (0.097 vs. 0.110); lower is better for CD evaluation metric.
 
Our model requires significantly fewer parameters than the baselines, since it performs joint regression using the shared feature map $\mathbf{X}_F$.
For example, the number of parameters used by our model is less than $1/12$ of the parameters used by LRM$_{\mathrm{w/ SAM \,+\, pix2gestalt}}$, as shown in Table~\ref{table:table_main_result}.

\paragraph{Qualitative comparison} 
Figure~\ref{fig:qualitative_comparison} presents single-view 3D shape reconstruction results on real-world images from Pix3D~\cite{Pix3D}.
When existing reconstruction models~\cite{hong2024lrm,huang2024zeroshape} utilize an off-the-shelf modal segmentation model~\cite{SAM}, they struggle to estimate the geometry for occluded object parts.
To enable occlusion-aware 3D reconstruction, they should incorporate an amodal completion model~\cite{pix2gestalt}. However, this approach often leads to error accumulation.
Even with well-completed amodal images, shape reconstruction errors may still occur (see LRM$_{\mathrm{w/ SAM \,+\, pix2gestalt}}$ in the left column of Figure~\ref{fig:qualitative_comparison}).
This might be due to compatibility issues with off-the-shelf models, as they were trained on different datasets.
In contrast, our model achieves promising results without the need to heavily pre-process input images using the off-the-shelf models.

\paragraph{Evaluation on non-occluded or occluded images only} 
In Table~\ref{table:table_non_occluded}, we compare evaluation results on non-occluded or occluded object images from the Pix3D dataset~\cite{Pix3D}, which consists of 42.6\% non-occluded and 57.4\% occluded images.
Under both scenarios, our end-to-end regression method substantially outperforms existing approaches~\cite{hong2024lrm,huang2024zeroshape} that depend on an off-the-shelf model~\cite{SAM}.

\begin{table}[!t]
    \centering
    \resizebox{\columnwidth}{!}{
        \begin{tabular}{l c c | c c}
        \toprule
        & \multicolumn{2}{c|}{Non-Occluded} & \multicolumn{2}{c}{Occluded}
        \\
        \cline{2-5}
        \raisebox{-0.15ex}{\!Model} & \raisebox{-0.15ex}{FS@$\tau$\;\!$\uparrow$} & 
        \raisebox{-0.15ex}{CD\,$\downarrow$} & \raisebox{-0.15ex}{FS@$\tau$\;\!$\uparrow$} & 
        \raisebox{-0.15ex}{CD\,$\downarrow$}
        \\
        \hline
        \raisebox{-0.15ex}{\!LRM$_{\mathrm{w/ SAM}}$} & \raisebox{-0.15ex}{33.5} & \raisebox{-0.15ex}{0.111} & \raisebox{-0.15ex}{29.1} & \raisebox{-0.15ex}{0.128}
        \\
        \!ZeroShape$_{\mathrm{w/ SAM}}$ & 34.6 & 0.106 & 30.2 & 0.123
        \\
        \hline
        \cellcolor{gray!7.5}\textbf{\!Ours \small{(category-agnostic)}}\!\!\! & \cellcolor{gray!7.5}\textbf{43.6} & \cellcolor{gray!7.5}\textbf{0.082} & \cellcolor{gray!7.5}\textbf{34.2} & \cellcolor{gray!7.5}\textbf{0.107}
        \\
        \bottomrule
    \end{tabular}}
    \vspace{-2mm}
    \caption{Quantitative evaluation exclusively on \textit{non-occluded} or \textit{occluded} object images from the Pix3D dataset~\cite{Pix3D}.}
    \vspace{-1.5mm}
    \label{table:table_non_occluded}
\end{table}

\begin{table}[!t]
    \centering
    \resizebox{\columnwidth}{!}{
        \begin{tabular}{l c c | c}
        \toprule
        & \multicolumn{3}{c}{Pix3D Evaluation}
        \\
        \cline{2-4}
        \raisebox{0.1ex}{\!Prompt} & \raisebox{-0.15ex}{\!FS@$\tau$\;\!$\uparrow$\!}  & \raisebox{-0.15ex}{\!FS@5$\tau$\;\!$\uparrow$\!} & \raisebox{-0.15ex}{CD\,$\downarrow$}
        \\
        \hline
        \raisebox{-0.15ex}{\!Category Agnostic\!\!} & \raisebox{-0.15ex}{38.2} & \raisebox{-0.15ex}{92.5} & \raisebox{-0.15ex}{0.097}
        \\
        \!Category Specific (w/ VLM)\; & \textbf{39.1} & \textbf{92.6} & \textbf{0.095}
        \\
        \bottomrule
    \end{tabular}}
    \vspace{-2mm}
    \caption{Pix3D~\cite{Pix3D} evaluation according to category prompts. 
    We incorporate category-specific priors by using a VLM~\cite{LLaMA-VID}.}
    \vspace{-1.5mm}
    \label{table:table_agnostic_specific}
\end{table}

\paragraph{Optional category-specific priors}
To evaluate the effect of incorporating such category priors, 
we compare results obtained using the category-agnostic prompt ``{object}'' and the category-specific prompt ``\texttt{[object]}'' (estimated by a VLM~\cite{LLaMA-VID}).
Table~\ref{table:table_agnostic_specific} shows that the priors improve our model's accuracy.
Please see the supplementary material for more analyses.

\begin{figure*}[t!]
    \centering
    \includegraphics[width=0.98\textwidth]{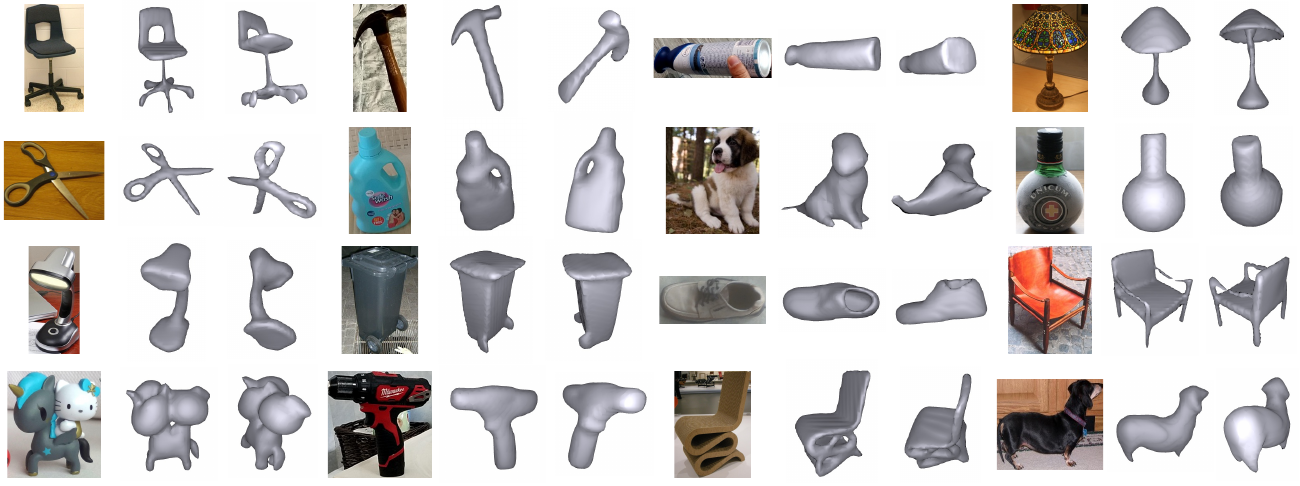}
    \vspace{-2mm}
    \caption{Single-view 3D shape reconstruction of diverse object shapes. 
    We qualitatively evaluate our model on real-world object images from ObjectNet~\cite{Objectnet}, OfficeHome~\cite{OfficeHomedataset}, and PACS~\cite{PACSdataset}.
    The proposed model achieves promising in-the-wild reconstruction results, which demonstrates that our regression-based model has effectively learned generalizable 3D shape priors.}
    \label{fig:fig_extensive_results}
    \vspace{-1.5mm}
\end{figure*}

\paragraph{Reconstruction of diverse objects} 
Figure~\ref{fig:fig_extensive_results} presents single-view 3D shape reconstruction results on real-world images from ObjectNet~\cite{Objectnet}, OfficeHome~\cite{OfficeHomedataset}, and PACS~\cite{PACSdataset}.
By leveraging learned priors, the proposed model can regress 3D shapes of various objects captured in real-world environments.

\begin{table}[!t]
    \centering
    \resizebox{\columnwidth}{!}{
        \begin{tabular}{l c c | c}
        \toprule
        & \multicolumn{3}{c}{Pix3D Evaluation}
        \\
        \cline{2-4}
        \!\raisebox{-0.15ex}{Model} & \raisebox{-0.15ex}{FS@$\tau$\;\!$\uparrow$}  & \raisebox{-0.15ex}{FS@5$\tau$\;\!$\uparrow$} & \raisebox{-0.15ex}{CD\,$\downarrow$}
        \\
        \hline
        \!\raisebox{-0.1ex}{\textit{w/o auxiliary depth}$^{*}$}  & \raisebox{-0.1ex}{35.9} & \raisebox{-0.1ex}{90.7} & \raisebox{-0.1ex}{0.105}
        \\
        \!\raisebox{-0.1ex}{\textit{w/o occluders}$^{*}$}  & \raisebox{-0.1ex}{37.7} & \raisebox{-0.1ex}{91.2} & \raisebox{-0.1ex}{0.102}
        \\
        \!\textit{w/o text prompt} & 38.0 & 91.5 & 0.101
        \\
        \hline
        \!Ours (category-specific)$^{*}$ & \textbf{39.6} & \textbf{92.8} & \textbf{0.094}
        \\
        \bottomrule
    \end{tabular}}
    \vspace{-1.75mm}
    \caption{Effects of key components. On Pix3D~\cite{Pix3D}, we analyze the accuracy of our model by ablating each component. $*$ denotes using ground-truth object names for category-specific prompts.}
    \label{table:table_ablation}
\end{table}

\subsection{Ablation Study}

\paragraph{Effect of auxiliary depth loss} 
In the absence of this loss,
we observe that predicted depth values for non-object regions are not precise, leading to worse results (see \textit{w/o auxiliary depth} in Table~\ref{table:table_ablation}).
This depth loss helps to accurately regress depth values across the entire image, which aids in distinguishing the salient object from its occluders and backgrounds based on depth. 
Moreover, the auxiliary loss prevents our model from overfitting to the training data.

\paragraph{Effect of simulating occluders} 
Without the Copy-Paste augmentation and the occluder mask estimation branch, our model struggles to reconstruct the full 3D shape when the salient object is occluded.
Although this variant estimates visible 3D shapes fairly well in the absence of occlusions, 
its overall accuracy is degraded on in-the-wild images from the Pix3D dataset~\cite{Pix3D} (see \textit{w/o occluders} in Table~\ref{table:table_ablation}).

\paragraph{Effect of text prompt} 
When we do not use text prompts, our model sometimes incorrectly splits a single object into visible and occluder regions based on depth. 
In addition, it struggles to handle noisy visible 3D shapes that contain background geometries by mistake.
As a result, its accuracy is diminished (see \textit{w/o text prompt} in Table~\ref{table:table_ablation}). 
\begin{table}[!t]
    \centering
    \resizebox{\columnwidth}{!}{
        \begin{tabular}{l | c | c c c}
        \toprule
        \!\!Model\! & \!\!Train\,GPUs\!\! & \!Feat.\,Res.\! & \!\!\!\!\!Train\,Res.\!\!\! & \!\!\!Inference\,Res.\!\!\!\!\!
        \\
        \hline
            \!\!LRM\! & \!128 A100\! & 32$\times$32 & \!\!\!\!\!\!128$\times$128$\times$128\!\!\!\! & 384$\times$384$\times$384\!\!
        \\
            \!\!Ours\! & \!4 RTX3090\! & 14$\times$14 & \!\!\!\!\!32$\times$32$\times$32\!\!\!  & 128$\times$128$\times$128\!\!
        \\
        \bottomrule
    \end{tabular}}
    \vspace{-1.75mm}
    \caption{Used resources and resolutions. \textsl{Feat.\,Res.} denotes image feature resolutions from DINOv2~\cite{DINOv2} or ResNet50~\cite{resnet}, while
    \textsl{Train\;\!/\;\!Inference Res.} indicate spatial resolutions.}
    \vspace{-0.25mm}
    \label{table:table_resolution}
\end{table}

\section{Limitations}
\paragraph{Lack of training resources}
As shown in Table~\ref{table:table_resolution}, resolutions 
differ significantly due to limited resources (\ie, 4 RTX3090 vs. 128 A100). Consequently, our reconstructed 3D shapes appear smoother than the shapes from LRM~\cite{hong2024lrm}.

\paragraph{Challenges with boundary-clipped objects}
Our model can handle occlusions, but it may struggle to estimate the geometry of object parts clipped by image boundaries.

\paragraph{Dataset ambiguity}
Ambiguity in the dataset may hinder our model's ability to distinguish the salient object from its occluders.
For example, some ``bed'' 3D objects only consist of bed frames, while others include pillows.

\paragraph{Unrealistic occluders}
Copy-Paste augmentations have proven effective~\cite{CopyPaste},
but they introduce a domain gap when compared to actual occluders in real-world environments.

\section{Conclusion}
To enable zero-shot single-view reconstruction in the wild, we presented (i) a unified regression model for occlusion-aware 3D shape reconstruction, and (ii) a scalable data synthesis pipeline for creating training dataset. Our approach achieved state-of-the-art accuracy on in-the-wild images. Unlike other existing methods, our method does not utilize off-the-shelf segmentation or amodal completion models, leading to significantly fewer model parameters.
\newpage
\paragraph{Acknowledgments}
This work was supported by ADD grant funded by the Korean government. 
T.-H. Oh and K. Youwang were partially supported by Institute of Information \& communications Technology Planning \& Evaluation (IITP) grant (No. RS-2024-00457882, National AI Research Lab Project; No. RS-2020-II200004, Development of Previsional Intelligence based on Long-term Visual Memory Network), and National Research Foundation of Korea (NRF) grant (No. RS-2024-00358135, Corner Vision: Learning to Look Around the Corner through Multi-modal Signals) funded by the Korean government (MSIT).

{
    \small \bibliographystyle{ieeenat_fullname}
    \bibliography{arXiv_main}
}

\clearpage
\setcounter{section}{0}
\setcounter{figure}{0}
\setcounter{table}{0}
\renewcommand{\thesection}{\Alph{section}}

\maketitlesupplementary

In this supplementary material, we provide more implementation details (Sec.~\ref{supp:supp_impl_details}), training and evaluation details (Sec.~\ref{supp:supp_training_details}), analyses (Sec.~\ref{supp:supp_more_analyses}),
discussion (Sec.~\ref{supp:supp_discussion}),
and broader impacts with ethics considerations (Sec.~\ref{supp:supp_ethics}),
which are not included in the main paper due to its space constraints.
\setcounter{figure}{0}
\setcounter{table}{0}
\renewcommand\thefigure{A\arabic{figure}}
\renewcommand\thetable{A\arabic{table}}

\section{Implementation Details}
\label{supp:supp_impl_details}

\subsection{Model Architecture}
\paragraph{Pixel-level regression}
We employ the backbone of DPT-Hybrid~\cite{DPT}, which is composed of $12$ transformer layers and $4$ fusion layers.
From the global feature map $\mathbf{X}_G$, we estimate 
parameters for camera intrinsics $K$ using $4$ convolutional layers, $2$ ReLU activation layers, an average pooling layer, and a linear layer.
From the fine-grained feature map $\mathbf{X}_F$,
we regress the depth map $\mathbf{M}_D$, visible mask $\mathbf{M}_V$, and occluder mask $\mathbf{M}_O$.
Specficially, the depth map $\mathbf{M}_D$ is regressed using $3$ convolutional layers, $2$ ReLU activation layers, and a bilinear interpolation layer.
The visible mask $\mathbf{M}_V$ and occluder mask $\mathbf{M}_O$ are regressed through $2$ convolutional layers, $1$ ReLU activation layer, and a bilinear interpolation layer, respectively.
The scaling factor $\gamma$ and shift factor $\beta$ used for the affine transformation are estimated via a linear layer and a SiLU activation layer.

\paragraph{3D point-wise regression} From the concatenated output of the visible 3D shape $\mathbf{S}_V$ and occluder mask $\mathbf{M}_O$, we extract features using the backbone of ResNet50~\cite{resnet}, augmented with additional $9$ convolutional layers and $4$ ReLU activation layers.
These extracted features serve as keys and values in $2$ cross-attention layers, where 
sine-cosine positional embeddings are incorporated to preserve spatial information. 
The cross-attention layers process queries constructed by the concatenation of point embeddings and text embeddings. 
The point embeddings are derived from 3D query points using $2$ linear layers,
while the text embeddings are obtained from CLIP text embeddings~\cite{radford2021clip} projected by a linear layer.
The outputs from the cross-attention layers are used to regress occupancy values of the 3D query points through $9$ linear layers with skip connections and $8$ Softplus activation layers.

\subsection{Data Synthesis}

The pseudocode for our data synthesis pipeline is outlined in Algorithm~\ref{alg:pseudocode}.

\begin{algorithm*}[t!]
    \begin{minipage}[c]{0.95\linewidth}
    \caption{\;\mbox{Pseudocode of our data synthesis pipeline}}
    \label{alg:pseudocode}
    \textbf{Requirement:}\;object renderer $\mathcal{R}(\cdot)$,\;guidance perturbator $\mathcal{P}(\cdot)$,\;random seed generator $\mathcal{G}(\cdot)$, \\
    \hspace*{\algorithmicindent}\hspace*{\algorithmicindent}\;\;\;\;\;\;\;\;\;\;\;\,diffusion model for object diversification $\mathtt{DM}_{\mathrm{obj}}(\cdot)$, diffusion model for background outpainting $\mathtt{DM}_{\mathrm{bg}}(\cdot)$ \\ 
    \textbf{Input:}\;3D objects $\{\mathcal{O}_{i}\}^{K}_{i=1}$,\;number of camera views for rendering 3D objects $\{N_{i}\}^{K}_{i=1}$, \\
    \hspace*{\algorithmicindent}\hspace*{\algorithmicindent}\;\!color list $\mathcal{L}_c$,\;material list $\mathcal{L}_m$,\;scene list $\mathcal{L}_s$,\;IoU filtering threshold $\kappa$ \\
    \textbf{Output:}
    $\langle \mathrm{3D\;object, \,camera\;parameters, \,depth\;map, \,mask, \,image} \rangle$\;\!--\;\!dataset
    \vspace{0.15mm}
    \begin{algorithmic}[1]
    \renewcommand{\alglinenumber}[1]{\LineNumberAlg{#1}}
        \Statex \commentcolor{\# $\mathcal{C}$: camera parameters,\; $\mathcal{D}$: depth map,\; $\mathcal{M}$: mask,\; $\mathcal{I}$: image}

        \vspace{1mm}
    
        \State 
        $\{\mathcal{C}_{i,j}\}_{i=1,j=1}^{K,N_{i}}, \{\mathcal{D}_{i,j}\}_{i=1,j=1}^{K,N_{i}}, \{\mathcal{M}_{i,j}\}_{i=1,j=1}^{K,N_{i}}, \{\mathcal{I}_{i,j}\}_{i=1,j=1}^{K,N_{i}} \leftarrow \mathcal{R}(\{\mathcal{O}_{i}\}^{K}_{i=1}, \{N_{i}\}^{K}_{i=1})$ 
        {\small \commentcolor{\Comment{render 3D objects}}}

        \vspace{0.75mm}
        
        \State 
        $\mathtt{dataset} \leftarrow [\,]$ 

        \For {$i=1,2,\ldots,K$}

            \For {$j=1,2,\ldots,N_{i}$} 

                \State
                $\mathtt{guide} \leftarrow \mathcal{P}(\mathcal{I}_{i,j})$
                {\small \commentcolor{\Comment{perturb an initial guidance}}}

                \While {$\mathtt{True}$} 
                
                    \State
                    $\mathtt{seed} \leftarrow \mathcal{G}()$
                    {\small \commentcolor{\Comment{set random seed}}}

                    \State
                    $\mathtt{[color]},\mathtt{[material]} \leftarrow \mathtt{sample}(\mathcal{L}_c, \mathcal{L}_m, \mathtt{seed})$
                    {\small \commentcolor{\Comment{randomly select words}}}

                    \State
                    $\mathtt{[object]} \leftarrow \mathtt{retrieve\_category}(\mathcal{O}_{i})$
                    {\small \commentcolor{\Comment{retrieve 3D object category}}}

                    \State
                    $\mathtt{txt} \leftarrow ``\mathtt{a}\;\mathtt{[color]}\;\mathtt{[material]}\;\mathtt{[object]}"$
                    
                    \State
                    $\mathtt{fg\_img} \leftarrow \mathtt{DM}_{\mathrm{obj}}(\mathcal{D}_{i,j}, \mathtt{guide},\mathtt{txt},\mathtt{seed})$
                    {\small \commentcolor{\Comment{simulate object appearance}}}

                    \If {\,$\mathtt{not}\;\mathtt{is\_filtered}(\mathtt{fg\_img},\mathcal{M}_{i,j},\kappa)$}
                    {\small \commentcolor{\Comment{filter an image with a threshold $\kappa$}}}
                    
                        \State
                        \textbf{break} %
                    \EndIf %
                \EndWhile

                \State
                $\mathtt{seed} \leftarrow \mathcal{G}()$
                {\small \commentcolor{\Comment{set random seed}}}

                \State
                $\mathtt{[scene]} \leftarrow \mathtt{sample}(\mathcal{L}_s, \mathtt{seed})$
                {\small \commentcolor{\Comment{randomly select a word}}}
                
                \State
                $\mathtt{txt} \leftarrow ``\mathtt{a}\;\mathtt{[object]}\;\mathtt{in}\;\mathtt{the}\;\mathtt{[scene]}"$
                
                \State
                $\mathtt{fg\_bg\_img} \leftarrow \mathtt{DM}_{\mathrm{bg}}(\mathcal{M}_{i,j}, \mathtt{fg\_img}, \mathtt{txt}, \mathtt{seed})$
                {\small \commentcolor{\Comment{simulate background}}}

                \State
                Add\;$(\mathcal{O}_{i}, \mathcal{C}_{i,j}, \mathcal{D}_{i,j}, \mathcal{M}_{i,j}, \mathtt{fg\_bg\_img})$\;to\;$\mathtt{dataset}$ %
            \EndFor %
        \EndFor
        \State
        \textbf{return} \,$\mathtt{dataset}$
    \end{algorithmic}
    \end{minipage}
\end{algorithm*} 

\paragraph{Rendering}
We utilize the 3D shape renderings provided by
ZeroShape~\cite{huang2024zeroshape}.
These renderings were produced using Blender~\cite{Blender} with various camera configurations.
Specifically, focal lengths were varied between $30$mm and $70$mm for a 35mm sensor size equivalent.
Camera distances and LookAt points were also randomized, with elevation angles ranging from
$5^{\circ}$ to $65^{\circ}$.
These renderings, produced at a resolution of $600 \times 600$ pixels, were cropped around the center of objects and resized to $224 \times 224$ pixels.

\paragraph{Object appearance diversification}
We utilize ControlNet~\cite{ControlNet} to simulate diverse visual variations in object appearances, excluding those with high-resolution texture maps (\eg, several objects from Objaverse~\cite{Objaverse}).
To be specific, variations are generated by a textual condition
``a \texttt{[color]} \texttt{[material]} \texttt{[object]}'', 
where \texttt{[color]} and \texttt{[material]} are randomly selected from pre-defined color list $\mathcal{L}_c$ and material list $\mathcal{L}_m$, respectively.
The color list $\mathcal{L}_c$ includes ``red'', ``pink'', ``orange'', ``yellow'', ``green'', ``blue'', ``purple'', ``brown'', ``white'', ``black'', ``gray'', and an empty string.
The material list $\mathcal{L}_m$ contains ``metal'', ``wood'', ``plastic'', ``ceramic'', ``stone'', ``rubber'', ``leather'', and an empty string.
To utilize more plausible and diverse textual conditions according to each object rendering,
one can leverage suggestions from LMMs~\cite{LLaMA-VID}.

\paragraph{Initial guidance}
As described in Section~\refclr{4.2} of the main paper, we leveraged initial guidance to reduce silhouette distortion of objects.
To be specific, when we utilized ControlNet~\cite{ControlNet},
we set $20$ steps in the DDIM sampler~\cite{DDIM} and injected the guidance at step $8$.
This guidance effectively forces the conditional diffusion model to precisely adhere to input spatial condition.

\paragraph{Filtering synthesized object images}
While the initial guidance substantially aids in preserving the object silhouette as specified in the spatial condition, minor disparities may still exist between the silhouettes in the synthesized images and the original silhouettes from the renderings.
To address this, we filter out images 
if the intersection-over-union (IoU) between the synthesized and original silhouettes is below $0.95$.
The silhouettes in the synthesized images are estimated by extracting the foreground objects in the images.
We can simply extract the foreground objects using a threshold,
as the initial guidance results in images that have nearly-white backgrounds.
Specifically, we convert each synthesized RGB image to grayscale, and then consider pixels with values between $250$ and $255$ as background.
This straightforward process allows us to accurately approximate the foreground silhouette in most cases.

\paragraph{Background diversification}
We simulate diverse backgrounds using an object-aware background outpainting model~\cite{salientobjbg}
with a textual condition ``a \texttt{[object]} in the \texttt{[scene]}'', 
where \texttt{[scene]} is randomly selected from pre-defined scene list $\mathcal{L}_s$.
As described in Section~\refclr{4.3} of the main paper, we use scene categories from~\cite{SUN,Places365} as the scene list $\mathcal{L}_s$,
which contains more than $700$ categories.
To utilize more plausible and diverse textual conditions according to each foreground object,
one can leverage suggestions from LMMs~\cite{LLaMA-VID}.
\setcounter{figure}{0}
\setcounter{table}{0}

\section{Training and Evaluation Details}
\label{supp:supp_training_details}
We initialize our model with ZeroShape~\cite{huang2024zeroshape} model weights for shared components such as DPT-Hybrid backbone~\cite{DPT} and cross-attention layers.
We first pre-train our pixel-level regression components for 10 epochs, using the Adam optimizer~\cite{Adam} with a learning rate of $10^{-5}$, a batch size of $80$, a weight decay of $0.05$, and momentum parameters of $(0.9, 0.95)$.
This process takes approximately $2$ days on $4$ RTX 3090 GPUs.
Then, we train our entire model for 15 epochs,
using the same optimizer with a learning rate of $10^{-5}$ for 3D point-wise regression components and $10^{-6}$ for pre-trained pixel-level regression components, a batch size of $80$, a weight decay of $0.05$ and momentum parameters of $(0.9, 0.95)$.
This process takes approximately $4$ days on $4$ RTX 3090 GPUs.

\renewcommand\thefigure{B\arabic{figure}}
\renewcommand\thetable{B\arabic{table}}

\begin{figure*}[t!]
    \centering
    \includegraphics[width=0.99\linewidth]
    {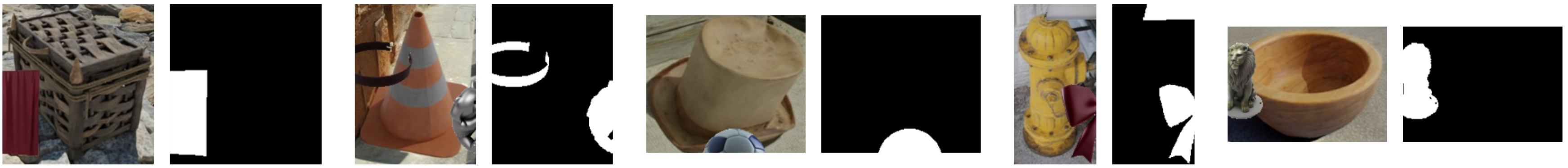}
    \vspace{-1.25mm}
    \caption{Examples of Copy-Paste augmentation. In this visualization, we provide augmented training samples with occluder masks.}
    \label{fig:supp_fig_copypaste}
    \vspace{1mm}
\end{figure*}

\setcounter{figure}{0}
\setcounter{table}{0}
\renewcommand\thefigure{C\arabic{figure}}
\renewcommand\thetable{C\arabic{table}}

\begin{figure*}[t!]
    \centering
    \includegraphics[width=0.99\linewidth]
    {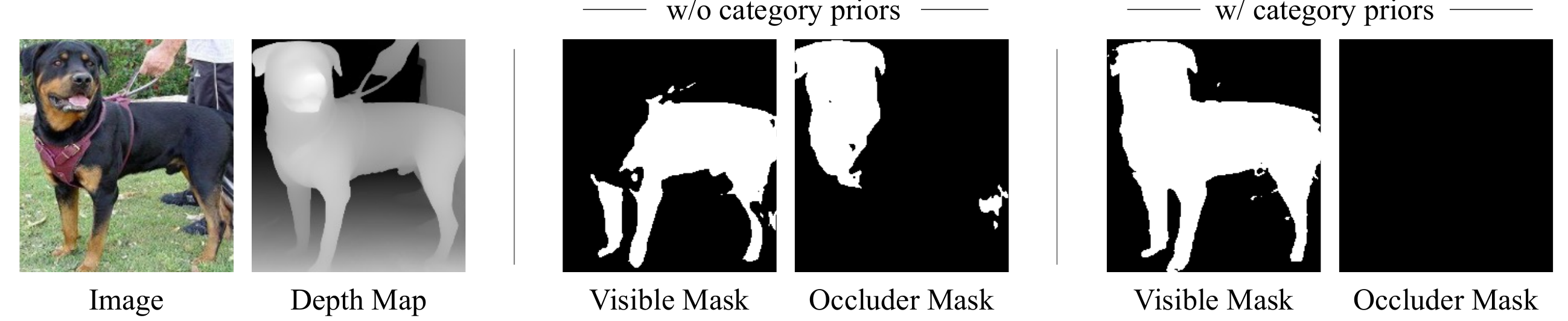}
    \vspace{-1.25mm}
    \caption{
    Effect of category priors on mask regression. Without utilizing the priors, the regression of visible and occluder masks appears to rely on depth values.
    By leveraging the priors, the regression is enhanced with semantic understanding.
    }
    \label{fig:supp_fig_category_vis_mask}
    \vspace{1mm}
\end{figure*}

\begin{figure*}[t!]
    \centering
    \includegraphics[width=0.96\linewidth]
    {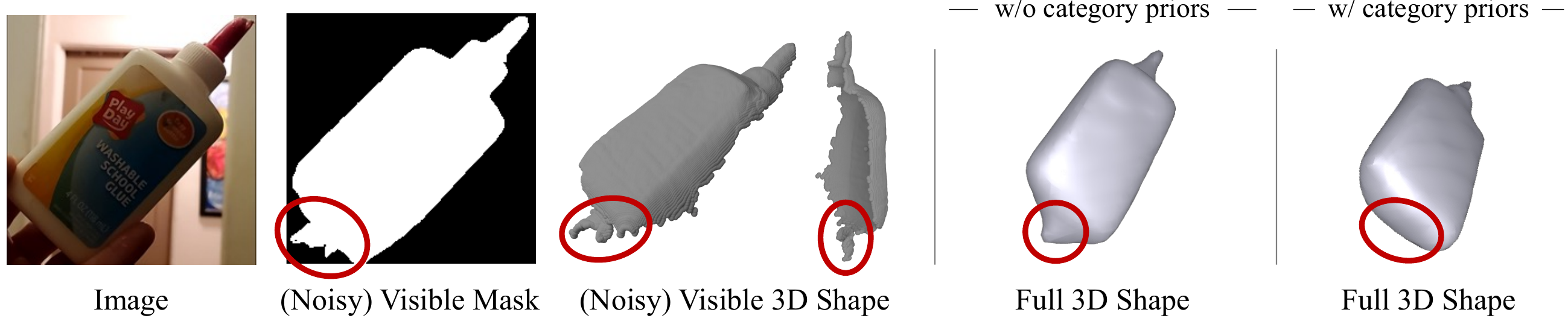}
    \vspace{-1.25mm}
    \caption{
    Effect of category priors on occupancy regression. Without utilizing the priors, it is challenging to distinguish the object from its background in the noisy visible 3D shape.
    By leveraging the priors, the regression is enhanced with learned 3D shape prior specific to the category, leading to more accurate results.
    We highlight regions with red circles.
    }
    \label{fig:supp_fig_category_occ}
    \vspace{3mm}
\end{figure*}

\paragraph{Loss coefficients}
Let $\mathcal{L}_c$ represent the camera intrinsics loss, $\mathcal{L}_d$ the depth loss using ground-truth depth values, $\mathcal{L}_d^{\mathrm{aux}}$ the auxiliary depth loss using depth values estimated from Depth Anything V2~\cite{DAV2},
$\mathcal{L}_m^{\mathrm{vis}}$ the visible mask loss, $\mathcal{L}_m^{\mathrm{occ}}$ the occluder mask loss, and $\mathcal{L}_o$ the occupancy loss.
The loss $\mathcal{L}$ used for training the pixel-level regression components is computed as follows:
\begingroup
\setlength{\abovedisplayskip}{-8pt}
\setlength{\belowdisplayskip}{8pt}
    \small
    \begin{align}
    \mathcal{L} = \lambda_c \mathcal{L}_c + \lambda_d \mathcal{L}_d + \lambda_d^{\mathrm{aux}} \mathcal{L}_d^{\mathrm{aux}} + \lambda_m^{\mathrm{vis}} \mathcal{L}_m^{\mathrm{vis}} + \lambda_m^{\mathrm{occ}} \mathcal{L}_m^{\mathrm{occ}},
    \end{align}
\endgroup
where $\lambda_c = 10, \lambda_d = \lambda_m^{\mathrm{vis}} =
\lambda_m^{\mathrm{occ}} = 1,$ and $\lambda_d^{\mathrm{aux}} = 0.1$.
The loss $\mathcal{L}'$ used for training our entire model is computed as follows:
\begingroup
\setlength{\abovedisplayskip}{5pt}
\setlength{\belowdisplayskip}{5pt}
    \begin{align}
     \mathcal{L}' = \mathcal{L} + \lambda_o \mathcal{L}_o,
    \end{align}
\endgroup
where $\lambda_o = 1$. 
To compute the occupancy loss $\mathcal{L}_o$,
we randomly sample $4096$ query points at every iteration.

\paragraph{Copy-Paste augmentation}
As shown in Figure~\ref{fig:supp_fig_copypaste}, we apply the augmentation by randomly selecting objects synthesized with focal lengths similar to the corresponding training sample.
In each training iteration, we randomly choose from 0 to 2 occluders,
resize them within a scale range of $0.4$ to $0.6$,
and put them onto the training sample.

\paragraph{Category priors}
We use CLIP text embeddings~\cite{radford2021clip} to learn category-specific priors with ground-truth object categories from Objaverse-LVIS~\cite{Objaverse} and ShapeNetCore.v2~\cite{ShapeNet}.

\paragraph{Evaluation}
For the quantitative evaluation of our model, we compute Chamfer Distance (CD) and F-Score (FS). Regarding FS@$\tau$,
we set the distance threshold $\tau$ to $0.05$. 
For the evaluation, we need to convert implicit 3D shapes into explicit meshes and then sample points from their surfaces.
To obtain explicit meshes,
we apply Marching Cubes~\cite{MarchingCubes} algorithm, using sampled values from a 
$128^3$ spatial grid.

\paragraph{Estimation of object category}
To incorporate category-specific priors,
we optionally estimate the object category of an object-centric image using a VLM~\cite{LLaMA-VID}.
We use the following prompt:
\textsl{What is the salient object in this image? The object should occupy the majority of the image. Please provide the category name of the salient object in the format: ``[object]", where ``[object]" is the specific category name (e.g., ``chair", ``bed", ``sofa", ``table").}

\section{More Analyses}
\label{supp:supp_more_analyses}

\setcounter{figure}{0}
\setcounter{table}{0}
\renewcommand\thefigure{D\arabic{figure}}
\renewcommand\thetable{D\arabic{table}}

\begin{figure*}[t!]
    \centering
    \includegraphics[width=0.99\linewidth]
    {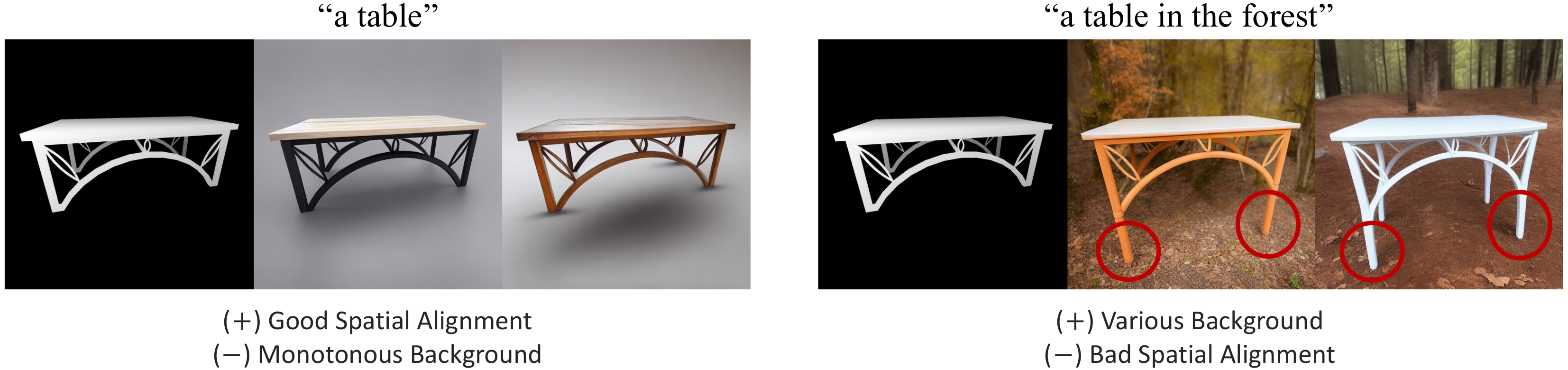}
    \vspace{-0.5mm}
    \caption{Observations from pre-trained conditional generative models~\cite{ControlNet,GLIGEN,T2I-Adapter,UniControl}.
    This phenomenon is mainly attributed to their pre-training procedure; 
    they were trained with depth maps which contain both foreground and background information.
    When we use depth maps rendered from 3D objects, synthesizing backgrounds violates input spatial conditions.}
    \label{fig:supp_fig_tradeoff}
    \vspace{3mm}
\end{figure*}

\paragraph{Category priors for regressing masks}
When category priors are not utilized, our model sometimes incorrectly splits a single entity (\eg, a dog) into visible region (\eg, the dog's body) and occluder region (\eg, the dog's head) based on depth values,
as shown in Figure~\ref{fig:supp_fig_category_vis_mask}.
We suspect this issue arises due to the lack of semantic understanding. 
To mitigate this, one may incorporate semantic priors by estimating object categories using a vision-language model.

\paragraph{Category priors for regressing occupancy values}
Visible 3D shapes may include background geometries due to noisy estimations of visible masks.
In such cases, it is challenging to accurately regress occupancy values for the corresponding objects. 
To be specific, distinguishing a salient object from its background is difficult,
because a visible 3D shape only contains the xyz-coordinates of each pixel (\ie, pixel-aligned point cloud)
without any supplemental visual cues (\eg, RGB color).
To address this issue, one may leverage category priors for regressing the occupancy values, as shown in Figure~\ref{fig:supp_fig_category_occ}.

\section{Discussion}
\label{supp:supp_discussion}

\noindent \textbf{Why synthesizing images in two steps?}
Our data synthesis pipeline first generates foreground objects and then outpaints their backgrounds. 
A more straightforward alternative would be to synthesize the entire image at once using conditional generative models such as ControlNet~\cite{ControlNet}.
However, as shown in Figure~\ref{fig:supp_fig_tradeoff}, 
when these models are forced to strictly follow the input spatial conditions, they often produce monotonous backgrounds due to the lack of background information in the conditions.
On the other hand, if we encourage the models to diversify backgrounds using textual conditions, they tend to create more varied backgrounds at the cost of violating the input spatial conditions. 
To resolve these challenges, we first diversify object appearances while adhering to the input spatial conditions, and then use an object-aware background outpainting model~\cite{salientobjbg},
specifically fine-tuned to prevent distortion of object silhouettes while generating the backgrounds.

\noindent \textbf{Various approaches for occlusion-aware reconstruction.}
Probabilistic methods (\eg, PT43D~\cite{PT43D}) are effective for handling heavily occluded objects by generating multiple plausible 3D shapes.
However, there are trade-offs between (i) accuracy and sample diversity~\cite{HuManiFlow}, and
(ii) accuracy and efficiency~\cite{gawlikowski2023survey}.
In comparison, regression-based methods can efficiently produce competitive results for small occlusions, but they become sub-optimal and often impractical when tackling highly occluded objects.

\setcounter{figure}{0}
\setcounter{table}{0}
\renewcommand\thefigure{E\arabic{figure}}
\renewcommand\thetable{E\arabic{table}}

\vspace{1mm}
\section{Broader Impacts \& Ethics Considerations}
\label{supp:supp_ethics}

Our data synthesis pipeline utilizes 3D object collections and conditional generative models.
To avoid conflicts,
one should carefully follow their usage rights, licenses and permissions.
Also, one should be aware that generative models might reflect biases inherent in their training data~\cite{esser2020note}, and object images in the training data might also be biased~\cite{vries2019does}.
Furthermore, one should keep in mind that generative models might expose their training data~\cite{carlini2021extracting}. 
To avoid data privacy issues,
one can use erasing methods~\cite{gandikota2023erasing,ni2023degenerationtuning} capable of removing unwanted concepts from generative models.

\end{document}